\documentclass[]{ceurart}
\usepackage{listings}
\usepackage{graphicx}
\usepackage{amsmath,amssymb}
\usepackage{algorithm}
\usepackage{algorithmic}
\usepackage{booktabs}
\usepackage{tikz}
\usetikzlibrary{arrows.meta,patterns,decorations.pathmorphing}

\begin{document}

%DO NOT CHANGE THE BLOCK BELOW
%%------------------------------------------
\copyrightyear{2026}
\copyrightclause{Copyright for this paper by its authors.
  Use permitted under Creative Commons License Attribution 4.0
  International (CC BY 4.0).}
\conference{Late-breaking work, Demos and Doctoral Consortium, colocated with the 4th World Conference on eXplainable Artificial Intelligence: July 01--03, 2026, Fortaleza, Brazil}
%%------------------------------------------

%%
%% Title
\title{Explaining Process Control Optimisation Recommendations via GradientSHAP and Implicit Differentiation}

%%
%% Authors and Affiliations
\author[1]{Paul Darm}[%
email=paul.darm@strath.ac.uk,
]

\author[3]{Cem Alpturk}[%
email=cem.alpturk@mail.weir,
]

\author[3]{Kenneth Ulrich}[%
email=kenneth.ulrich@mail.weir,
]

\author[2]{William Duncan}[%
email=william.duncan@mail.weir,
]

\author[1]{Ali Anwar}[%
email=ali.anwar@strath.ac.uk,
]

\author[1]{Annalisa Riccardi}[%
email=annalisa.riccardi@strath.ac.uk,
]

\address[1]{University of Strathclyde, Glasgow, United Kingdom}
\address[2]{Weir, Glasgow, United Kingdom}
\address[3]{Weir, Malm\"o, Sweden}

%%
%% Abstract
\begin{abstract}
Automated optimisation is increasingly adopted in industrial processes, yet a trust gap persists between engineers who design these algorithms and operators who must act on their recommendations. Explainable AI methods like SHAP (SHapley Additive exPlanations) have transformed interpretability for machine learning predictions; optimisation outputs could benefit from similar techniques. We present an approach that integrates Implicit Function Theorem (IFT) based sensitivity analysis with SHAP attribution and narrative generation via Large Language Models (LLM), producing explanations tailored for operators. Our approach leverages IFT to compute exact parameter sensitivities $\partial p^*/\partial x$ from the optimality conditions, enabling efficient GradientSHAP computation. For an industrial High Pressure Grinding Roll (HPGR) control optimisation problem with 22 features, we achieve equivalent SHAP attributions (correlation $>$0.99 with KernelSHAP) with over 40$\times$ speedup, enabling real-time natural language explanations. We validate on industrial scenarios and present feedback from domain experts on generated explanations.
\end{abstract}

%%
%% Keywords
\begin{keywords}
Explainable AI \sep
Explainable Optimisation \sep
Implicit Function Theorem \sep
SHAP \sep
Sensitivity Analysis \sep
LLM \sep
HPGR
\end{keywords}

%%
%% Build the first part of the formatted document
\maketitle

%
% 1. INTRODUCTION
%
\section{Introduction}

Modern industrial processes increasingly rely on mathematical optimisation for operational control, offering significant economic and efficiency benefits. However, seamless adoption faces a critical barrier: operators often distrust automated recommendations they cannot understand~\cite{liu2023increasing}. In high-stakes applications such as mineral processing, where safety and equipment constraints are paramount, explanations can ensure that optimized parameters lead to more efficient and safe operation by enabling operators to validate recommendations before implementation.Explainable Artificial Intelligence (XAI) has made significant progress in interpreting machine learning model predictions~\cite{lundberg2017unified,ribeiro2016should}. We apply a similar philosophy to explain optimisation algorithm outputs, answering operator questions like: \textit{``Why did the optimiser recommend higher pressure today?''} or \textit{``What input changes are driving this speed recommendation?''} by providing attributions of input features in form of approximated SHAP values to an LLM. The LLM then generates a narrative based on the SHAP values to explain why the parameters settings have changed.

This paper makes the following contributions:

\begin{enumerate}
    \item \textit{GradientSHAP for optimisation:} We apply GradientSHAP to explain optimisation algorithm outputs efficiently by deriving the required gradients $\partial p^*/\partial x$ from the Implicit Function Theorem.
    \item \textit{SHAP-based explanations:} We use the resulting attributions to generate natural language explanations for optimised parameters for an HPGR process for operators, validating it through feedback from domain experts.
\end{enumerate}

%
% 2. RELATED WORK
%
\section{Related Work}
\label{sec:related}

\textit{Explainability for Machine Learning.} The challenge of interpreting black-box model predictions has driven extensive research in explainable AI. 
SHAP~\cite{lundberg2017unified} provides game-theoretic feature attributions with properties like local accuracy and consistency. GradientSHAP~\cite{sundararajan2017axiomatic} computes attributions by integrating gradients along paths from baseline to instance, offering efficiency for differentiable models. LIME~\cite{ribeiro2016should} fits local surrogate models via perturbation sampling. Henkel et al.~\cite{henkel2024interpretable} applied SHAP to neural networks within model predictive control for model transparency. Martens et al.~\cite{martens2024tell} demonstrated that LLMs can convert numerical attributions into human-readable narratives. %\noindent\par

\textit{Explainability in Optimisation.} Unlike XAI for ML, explainability in optimisation remains nascent~\cite{bock2024explainable}. Korikov et al.~\cite{korikov2021counterfactual} explored counterfactual explanations via inverse optimisation, identifying minimal changes to make a desired solution optimal. Biemans et al.~\cite{biemans2025explainable} presented a LIME-inspired sampling approach to generate feature importance attributions for supply chain scheduling, requiring $N$ optimisation solves per explanation. An LLM was then used to generate narratives for specific supply chain scheduling decisions. %Our work differs by \textit{exploiting} the mathematical structure of converged optimisation problems: applying IFT to optimality conditions yields exact sensitivities with a single solve plus gradient computation.

\textit{Differentiable Optimisation.} A growing body of work treats optimisation problems as differentiable components within larger learning systems, computing gradients of optimal solutions via the implicit function theorem applied to optimality or KKT conditions~\cite{amos2017optnet, agrawal2019differentiable, xu2023revisiting}. These methods enable end-to-end learning of cost functions, constraints, or dynamics from data. %Our work applies the same underlying technique to a different end: rather than embedding an optimiser inside a trainable model, we use implicit differentiation to make an optimiser's recommendations explainable.

Our work sits at the intersection of these three lines: we apply implicit differentiation, the same technique used to embed optimisers as trainable layers, to instead compute exact sensitivities with a single solve, which we then use to generate SHAP-based narative explanations of optimiser recommendations. 

%
% 3. PROBLEM FORMULATION
%
\section{Problem Formulation}
\label{sec:problem}

\subsection{HPGR Parameter Optimisation}

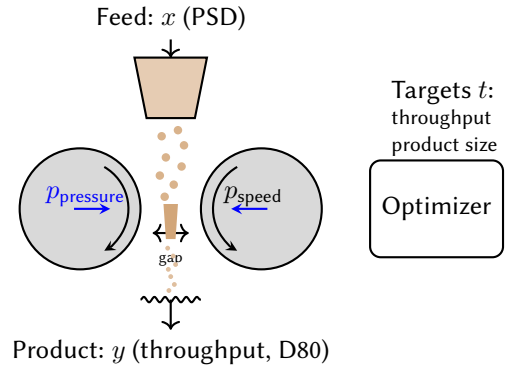
\begin{figure}[htbp]
\centering
\begin{tikzpicture}[scale=0.8]
    % Left roll
    \draw[thick, fill=gray!30] (-1.5,0) circle (1);
    \draw[thick, ->, >=stealth] (-1.5,0) ++(60:0.8) arc (60:-60:0.8);

    % Right roll
    \draw[thick, fill=gray!30] (1.5,0) circle (1);
    \draw[thick, ->, >=stealth] (1.5,0) ++(120:0.8) arc (120:240:0.8);

    % Hydraulic pressure arrows
    \draw[thick, ->, >=stealth, blue] (-1.6,0) -- (-1,0);
    \draw[thick, ->, >=stealth, blue] (1.6,0) -- (1,0);
    \node[blue, left] at (-0.6,0.2) {$p_{\text{pressure}}$};

    % Speed labels
    \node[right] at (0.73,0.2) {\small $p_{\text{speed}}$};

    % Feed (top)
    \draw[thick, fill=brown!40] (-0.6,2.5) -- (0.6,2.5) -- (0.4,1.5) -- (-0.4,1.5) -- cycle;
    \draw[thick, ->] (0,2.8) -- (0,2.5);
    \node[above] at (0,2.8) {\small Feed: $x$ (PSD)};

    % Product (bottom)
    \draw[thick, decorate, decoration={snake, amplitude=1pt, segment length=4pt}]
        (-0.5,-1.5) -- (0.5,-1.5);
    \draw[thick, ->] (0,-1.5) -- (0,-2.0);
    \node[below] at (0,-2.0) {\small Product: $y$ (throughput, D80)};

    % Gap
    \draw[<->, thick] (-0.3,-0.4) -- (0.3,-0.4);
    \node[below] at (0,-0.6) {\tiny gap};

    % Targets annotation
    \node[right, align=left] at (3.5,1.5) {\small Targets $t$:\\[-2pt] \scriptsize throughput\\[-2pt] \scriptsize product size};

    % Optimizer box
    \draw[rounded corners, thick] (3.3,-0.8) rectangle (5.5,0.8);
    \node at (4.4,0) {\small Optimizer};

    % Material flow - coarse particles entering between rolls
    \foreach \x/\y in {-0.15/1.2, 0.12/1.3, 0.22/1.05, -0.08/0.85, 0.18/0.7, -0.18/0.55, 0.05/0.4, -0.1/0.2} {
        \fill[brown!60] (\x,\y) circle (0.07);
    }
    % Compressed material in nip zone
    \fill[brown!70] (-0.12,0.08) -- (0.12,0.08) -- (0.08,-0.5) -- (-0.08,-0.5) -- cycle;
    % Finer discharged particles below gap
    \foreach \x/\y in {-0.05/-0.65, 0.04/-0.78, 0.08/-0.92, -0.04/-1.0, 0.0/-1.12, 0.06/-1.25, -0.03/-1.35} {
        \fill[brown!50] (\x,\y) circle (0.04);
    }

\end{tikzpicture}
\caption{HPGR simplified schematic. Two counter-rotating rolls crush feed material under hydraulic pressure. The optimiser adjusts pressure and speed to minimise deviation from production targets given current feed properties.}
\label{fig:hpgr_schematic}
\end{figure}

High Pressure Grinding Rolls (HPGR) are critical comminution equipment in mineral processing~\cite{dundar2013hpgr}, crushing ore between two counter-rotating rolls under high hydraulic pressure (Fig.~\ref{fig:hpgr_schematic}). Operators must continuously adjust two control parameters, operating pressure ($p_{\text{pressure}}$) and peripheral roll speed ($p_{\text{speed}}$), to achieve production targets while respecting equipment constraints.

The parameter optimisation problem is formulated as:

\begin{equation}
\min_{p \in \mathbb{R}^2} L(p, x, t)
\label{eq:hpgr_opt}
\end{equation}

where:
\begin{itemize}
    \item $p = [p_{\text{pressure}}, p_{\text{speed}}]^T$ are control parameters
    \item $x \in \mathbb{R}^n$ is the process input  particle size distribution (PSD), a 20-dimensional array of cumulative passing percentages at standard sieve sizes
    \item $t \in \mathbb{R}^m$ are production targets (throughput, product fineness)
    \item $L(p, x, t)$ is a loss function measuring deviation from targets
\end{itemize}

In practice, operating bounds constrain the feasible region, but for the scenarios in this paper, optimal solutions lie in the interior, allowing us to apply unconstrained sensitivity analysis. Extension to constraint boundaries is addressed in future work.

The loss function $L$ uses a physics-based population balance model~\cite{dundar2013hpgr} with data-fitted breakage parameters, implemented in JAX and fully differentiable, that predicts throughput and product fineness (D80---the particle size below which 80\% of the product passes). The loss measures weighted deviations from targets:
$$L(p, x, t) = w_1 (\hat{y}_{\text{throughput}} - t_{\text{throughput}})^2 + w_2 (\hat{y}_{\text{D80}} - t_{\text{D80}})^2$$

\subsection{Explanation Task}

Given two operating points:
\begin{itemize}
    \item \textit{Baseline:} inputs $x_0$, targets $t_0$, optimal parameters $p_0^* = \arg\min_p L(p, x_0, t_0)$
    \item \textit{Instance:} inputs $x_1$, targets $t_1$, optimal parameters $p_1^* = \arg\min_p L(p, x_1, t_1)$
\end{itemize}

We seek to explain: \textit{Why did optimal parameters change from $p_0^*$ to $p_1^*$?}

Specifically, we want an attribution breakdown:
$$\Delta p^* = p_1^* - p_0^* = \sum_{i} \phi_i^{(x)} + \sum_{j} \phi_j^{(t)}$$
where $\phi_i^{(x)}$ represents the contribution of input $x_i$ and $\phi_j^{(t)}$ represents the contribution of target $t_j$.

%
% 4. METHODOLOGY
%
\section{Methodology}
\label{sec:method}

\subsection{IFT for Sensitivity Computation}

Let $s = (x, t)$ denote the combined vector of process inputs and production targets. The optimal parameters $p^*(s)$ are implicitly defined by the optimality conditions. For unconstrained optimisation, first-order optimality requires:
$$\nabla_p L(p^*, s) = 0$$

By the Implicit Function Theorem, if the Hessian $H = \nabla_p^2 L(p^*, s)$ is non-singular, then $p^*(s)$ is locally differentiable:

\begin{equation}
\frac{\partial p^*}{\partial s} = -H^{-1} \cdot \frac{\partial^2 L}{\partial p \partial s}\bigg|_{(p^*, s)}
\label{eq:ift_unconstrained}
\end{equation}

See Dontchev and Rockafellar~\cite{dontchev2014implicit} for the general theory of implicit differentiation in optimisation.
The Hessian is computed once; scalar variables (moisture, targets) are batched together, while array-valued inputs (PSD) require a separate batch. Automatic differentiation via JAX~\cite{blondel2022efficient} computes these efficiently, adding negligible overhead compared to the optimisation solve.

\subsection{GradientSHAP for Optimisation}

With sensitivities $\partial p^*/\partial s$ available via IFT, we apply GradientSHAP~\cite{lundberg2017unified} to compute SHAP values. GradientSHAP integrates gradients along a path from a baseline to the instance of interest. For a differentiable function $h: \mathbb{R}^d \to \mathbb{R}$, the SHAP value for feature $i$ is:

\begin{equation}
\phi_i = (s_i - s_i^{\text{baseline}}) \int_{\alpha=0}^1 \frac{\partial h}{\partial s_i}\bigg|_{s^{\text{baseline}} + \alpha(s - s^{\text{baseline}})} d\alpha
\label{eq:gradshap}
\end{equation}

In our case, $h$ is the optimal parameter function $p^*(s)$, mapping inputs and targets to optimal operating parameters. Unlike neural networks where gradients are readily available, computing $\partial p^*/\partial s$ for an optimiser's output is not directly possible. IFT enables this by differentiating through the optimality conditions (Eq.~\ref{eq:ift_unconstrained}), which in turn enables GradientSHAP with theoretical efficiency gains over black-box methods such as KernelSHAP that require repeated optimisation solves.

\subsection{Narrative Generation with LLMs}

We use OpenAI (GPT-5) to convert SHAP attributions into operator-friendly narratives. The prompt provides domain knowledge about HPGR physics and the numerical SHAP breakdown; the LLM then interprets these values in operationally meaningful terms. Unlike surrogate-model approaches, GradientSHAP attributions derive directly from gradients of the actual optimiser, capturing exactly how the optimal output would change for small perturbations of each input, providing a principled foundation for faithful explanations. Crucially, the LLM acts as a translator of numerical attributions, not a reasoner about physics: the prompt grounds it in SHAP values computed from the actual optimiser, recommended parameter values come from the optimiser rather than the LLM, and operators validate all recommendations before implementation.

%
% 5. Results
\section{Results}
\label{sec:results}

\subsection{Sample Efficiency and Computational Performance}

\begin{figure}[htbp]
    \centering
    \includegraphics[width=0.8\textwidth]{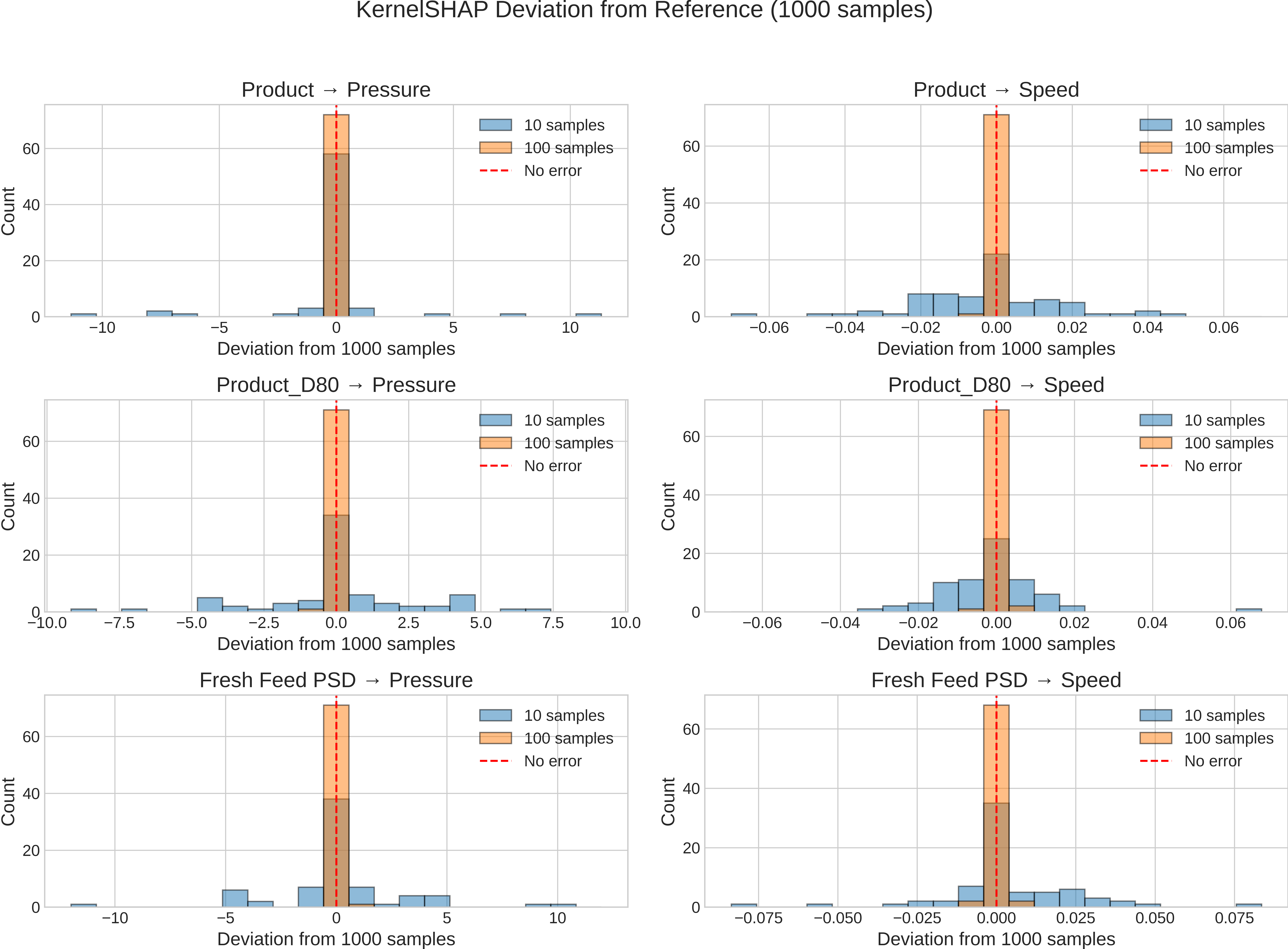}
    \caption{KernelSHAP convergence: deviation from 1000-sample reference. Blue: 10 samples (wide spread); Orange: 100 samples (concentrated near zero). KernelSHAP requires approximately 100 samples for stable results.}
    \label{fig:kernelshap_convergence}
\end{figure}

\begin{figure}[htbp]
    \centering
    \includegraphics[width=0.8\textwidth]{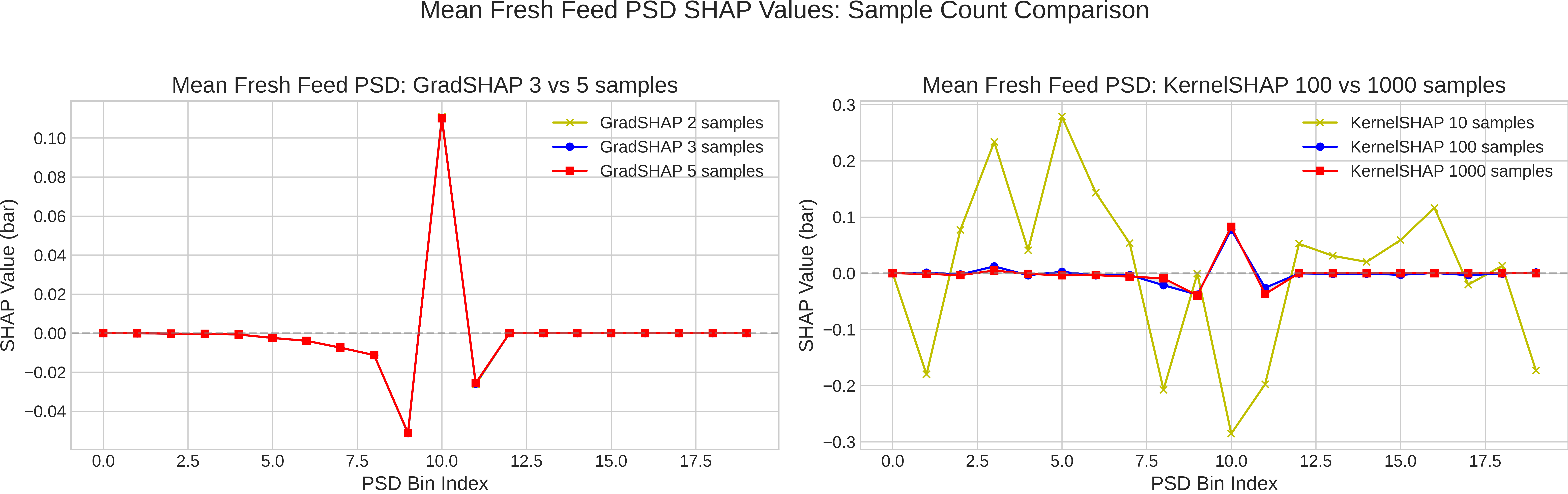}
    \caption{Sample count comparison for PSD SHAP values. Left: GradientSHAP shows rapid convergence with few samples. Right: KernelSHAP requires many more samples for stability.}
    \label{fig:sample_comparison}
\end{figure}

We examine convergence behavior of each approach. Figure~\ref{fig:kernelshap_convergence} shows KernelSHAP requires approximately 100 samples to reduce variance, while Figure~\ref{fig:sample_comparison} demonstrates GradientSHAP produces stable results with as few as 2--3 path integration samples. This difference in sample efficiency yields over 40$\times$ speedup (Table~\ref{tab:timing}). GradientSHAP timing is relatively stable across sample counts because the expensive JAX compilation of the Hessian computation occurs once and is amortised across path points.

\begin{table}[htbp]
\centering
\caption{Timing comparison between KernelSHAP and GradientSHAP. Speedup is relative to converged KernelSHAP (1000 samples).}
\label{tab:timing}
\begin{tabular}{llrr}
\toprule
\textit{Method} & \textit{Samples} & \textit{Mean Time (s)} & \textit{Speedup} \\
\midrule
KernelSHAP & 10 & 20.5 & -- \\
KernelSHAP & 100 & 38.1 & -- \\
KernelSHAP & 1000 & 362.9 & 1.0$\times$ (ref) \\
\midrule
GradientSHAP & 2 & 7.7 & 47.1$\times$ \\
GradientSHAP & 3 & 8.0 & 45.3$\times$ \\
GradientSHAP & 5 & 8.6 & 42.2$\times$ \\
\bottomrule
\end{tabular}
\end{table}

\subsection{GradientSHAP Validation}

\begin{figure}[htbp]
    \centering
    \includegraphics[width=0.7\textwidth]{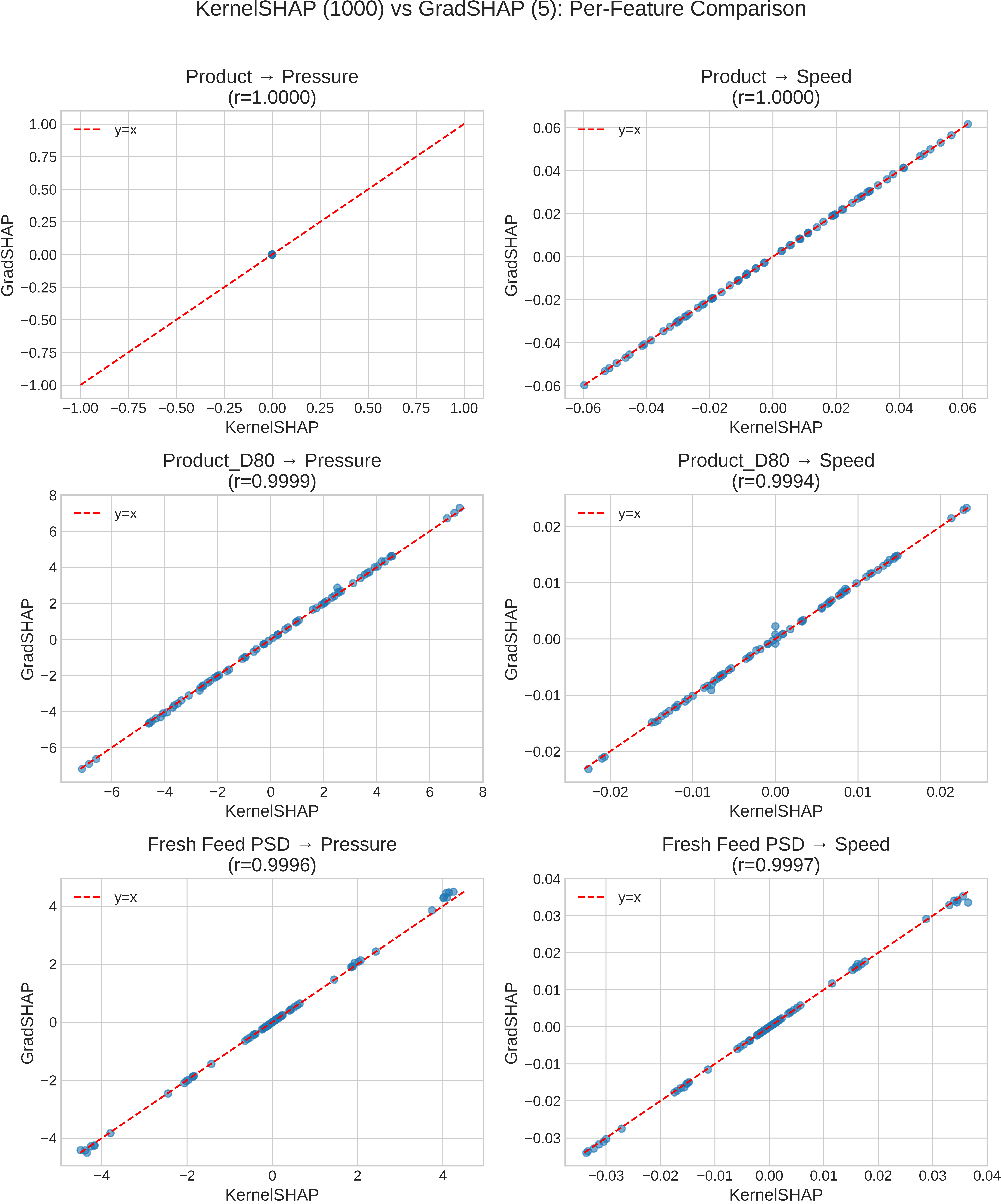}
    \caption{Correlation between converged KernelSHAP and GradientSHAP for scalar input features. Correlation exceeds 0.99 for all features except product target (pressure-independent), validating that GradientSHAP correctly approximates SHAP values.}
    \label{fig:correlation_scalar}
\end{figure}

Having established that GradientSHAP converges with few samples, we now validate that it produces correct SHAP values by comparing against converged KernelSHAP~\cite{lundberg2017unified} across 72 baseline-instance pairs. Figure~\ref{fig:correlation_scalar} shows correlation plots for scalar input features, demonstrating strong agreement ($r > 0.99$) between converged KernelSHAP and GradientSHAP with few path integration samples.

\begin{figure}[htbp]
    \centering
    \includegraphics[width=0.7\textwidth]{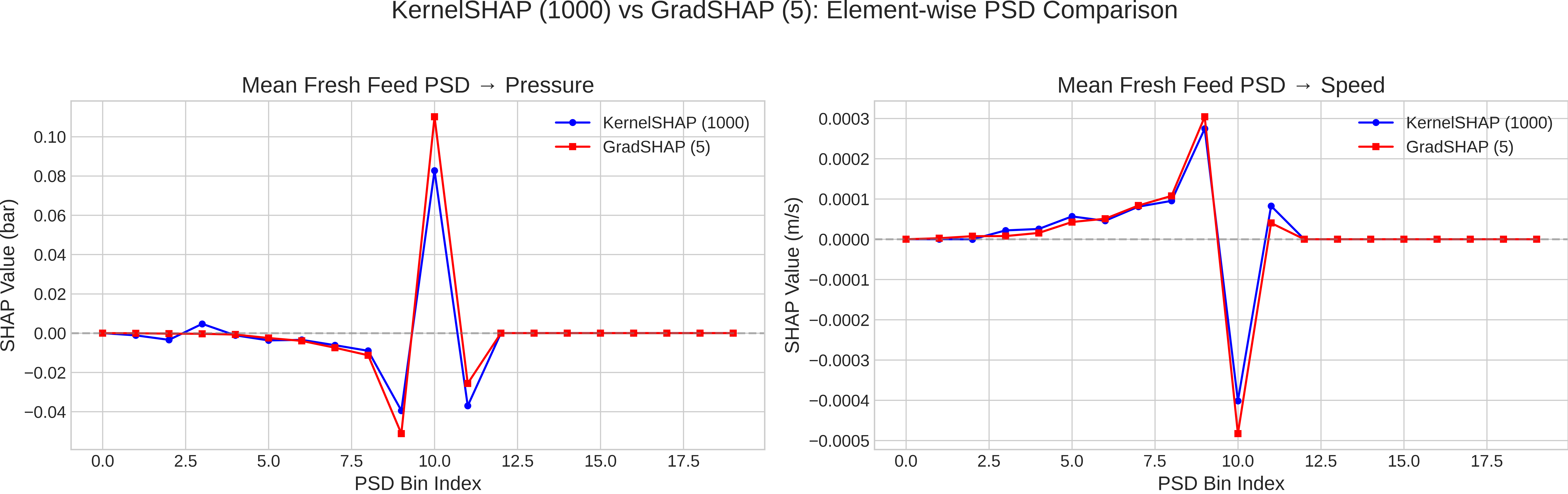}
    \caption{Mean element-wise SHAP values for fresh feed PSD, averaged over the dataset. Left: pressure output; Right: speed output. KernelSHAP (1000 samples, blue) and GradientSHAP (5 samples, red) show excellent agreement across all 20 PSD bins.}
    \label{fig:psd_comparison}
\end{figure}

For array-valued inputs, Figure~\ref{fig:psd_comparison} compares element-wise SHAP values averaged over the dataset for the 20-dimensional particle size distribution (PSD). Both methods produce nearly identical attribution profiles, with characteristic peaks around bins 7--10 indicating that mid-range particle sizes have the strongest influence on optimal parameters. PSD changes primarily affect pressure recommendations (SHAP values $\pm$0.10 bar) with weaker effects on speed ($\pm$0.2\%). The close agreement confirms that GradientSHAP correctly approximates KernelSHAP for both scalar and array-valued inputs.

\section{Explanations}
\label{sec:explanations}

The SHAP attributions computed via IFT-based GradientSHAP were used to generate natural language explanations for multiple test scenarios. We gathered informal qualitative feedback from three domain experts (two process engineers and one HPGR specialist), who evaluated the explanations for clarity and practical utility. Overall, they found the generated explanations helpful and accurate in capturing why the optimiser recommended specific parameter changes; a formal user study is planned as future work.

However, the evaluation also revealed important design principles for effective explanations. First, experts consistently skipped detailed attribution breakdowns and sought the final recommendation immediately; explanations should therefore lead with a one-sentence actionable summary before presenting supporting details. In addition, returning raw SHAP values together with the explanation for transparency were unhelpful to practitioners unfamiliar with the method; as one reviewer noted, \textit{`As someone with no
background in SHAP, the section 'What These SHAP Values Mean isn't clear to me.''}. So raw SHAP values were omitted.  Finally, technical precision of the algorithm output and thus the explanation is not useful for practitioners which preferred percentages over raw units (``73\% of max speed''), rounded values (``165 bar'' not ``166.8 bar'').

Based on this feedback, the explanation generation was adapted to incorporate these principles. The following scenario demonstrates an explanation that follows these principles.

\subsection{Scenario: Reduced Throughput with Coarser Feed*}

The following is a summarisation of one generated explanation by our pipeline:

\textit{Operating Point Change:}
\begin{itemize}
    \item Throughput target: 280 $\to$ 270 t/h ($-$10 t/h)
    \item Product fineness (D80) target: 1450 $\to$ 1460 $\mu$m (+10 $\mu$m, slightly coarser)
    \item Feed PSD: coarser across all size classes (1--2\% fewer fines near screen cut)
\end{itemize}

\textit{Recommended Adjustments:}
\begin{itemize}
    \item Pressure: 125 $\to$ 122 bar ($-$3 bar); 75\% $\to$ 70\% of range
    \item Speed: 88\% $\to$ 87\% of max ($-$1\%); 65\% $\to$ 64\% of range
\end{itemize}

\textit{Generated Explanation (condensed):}

\begin{quote}
\textit{One-sentence summary:} Because the operator wants lower throughput and a slightly coarser product, the system recommends backing off both pressure and roll speed, even though the feed itself is coarser.

\textit{Pressure ($-$3 bar):} The operator allows the product to be about 10 $\mu$m coarser, so the HPGR does not need to crush as hard, this alone explains about 3.5 bar of pressure reduction. The coarser feed (fewer fines near the screen cut) partially offsets this, pushing pressure up by about 0.5 bar. Net result: pressure decreases by 3 bar.

\textit{Speed ($-$1\%):} The operator explicitly wants less tonnage. To achieve that, the rolls should slow down slightly. The lower throughput request is the clear reason for reducing speed by about 1\%.

Both changes are small, stable, and well within operating limits.\footnote{\textit{*Values modified for confidentiality; relative relationships preserved.}}
\end{quote}

This explanation demonstrates: (1) leading with a one-sentence summary, (2) explicit baseline-to-new context, (3) SHAP values translated to operational meaning (``this alone explains about 3.5 bar''), and (4) operator-friendly units (percentages, rounded values, screen cut reference).
%
% 7. CONCLUSION
%
\section{Conclusion}
\label{sec:conclusion}

This paper presented the first application of GradientSHAP to explain optimisation recommendations, leveraging the Implicit Function Theorem for efficient sensitivity computation. By exploiting the mathematical structure of optimisation rather than treating it as a black box, we achieve over 40$\times$ speedup compared to KernelSHAP while maintaining correlation $>$0.99 in SHAP attributions. Integration with LLM-based narrative generation provides a complete pipeline from optimisation outputs to operator-friendly explanations, validated through feedback from process engineers and plant operators.

A limitation of the current approach is that it applies only to unconstrained optimisation, where solutions lie in the interior of the feasible region. Extension to constrained problems requires differentiating through the full Karush–Kuhn–Tucker (KKT) system to handle changes in the active set~\cite{dontchev2014implicit}. In addition, both the loss function and process model must be differentiable. SHAP values are also local attributions which means that the generated explanations are specific to a baseline-instance pair. While demonstrated on HPGR, the methodology applies broadly to differentiable optimisation in process control.%, scheduling, and energy systems. 
Future work includes expanding the methodology to constrained optimisation and formal user studies of the generated explanations. Code is available on request due to industrial confidentiality.

%%
%% Acknowledgments
\begin{acknowledgments}
We thank Chris Zerr, Dr Renato Oliveira, and Monica Botha Hansson for their feedback regarding the generated explanations.
\end{acknowledgments}

%%------------------------------------------
%%-----THE SECTION BELOW IS COMPULSORY------
%%------------------------------------------
\section*{Declaration on Generative AI}
During the preparation of this work, the authors used GPT-5 in order to generate natural language explanations from SHAP attributions as described in the methodology. Claude Opus 4.6 was used to correct grammar and spelling mistakes for the main text. After using Claude, the authors reviewed and edited the content as needed and takes full responsibility for the publication's content.
%%------------------------------------------

%%
%% Bibliography
\bibliography{references}

\end{document}